\documentclass{ieeeaccess}

\usepackage[utf8]{inputenc}
\usepackage{cite}
\usepackage[pdftex]{graphicx}
\usepackage{amsfonts, amsmath}
\usepackage{url}
\usepackage{algorithm}
\usepackage{algorithmic}
\usepackage{multirow}
\usepackage{stfloats}
\usepackage[kerning=true]{microtype}
\usepackage[all]{nowidow}
\usepackage{caption}

\newcommand{\R}{\mathbb R}

\newcommand{\dimension}{k}
\newcommand{\hDim}{m}

\newcommand{\Fun}{F}
\newcommand{\Surr}{S}
\newcommand{\Rel}{R}
\newcommand{\Dec}{D}
\newcommand{\Predictor}{P}
\newcommand{\hist}{h}

\newcommand{\fSupp}{\R^\dimension}
\newcommand{\hSupp}{(\R^{\dimension + \hDim})^\ast}
\newcommand{\fullSupp}{\fSupp \times \hSupp}

\newcommand{\MM}{\mathit{MM}}
\newcommand{\relevance}{\mathit{relevance}}
\newcommand{\threshold}{\Theta}

\DeclareMathOperator*{\argmin}{argmin}
\DeclareMathOperator*{\avg}{avg}

\newlength{\xfigwd}
\DeclareCaptionFont{ieeeblue}{\color{accessblue}}
\DeclareCaptionLabelFormat{myformat}{\figcapfont{\textbf{#1}\textbf{#2}}}
\begin{document}

\history{Date of publication xxxx 00, 0000, date of current version xxxx 00, 0000.}
\doi{10.1109/ACCESS.20XX.DOI}

\title{Meta-Model Framework for Surrogate-Based Parameter Estimation in Dynamical Systems}

\author{\uppercase{\v{Z}iga~Luk\v{s}i\v{c}}\authorrefmark{1}, \uppercase{Jovan~Tanevski}\authorrefmark{2}, \uppercase{Sa\v{s}o~D\v{z}eroski}\authorrefmark{2}, and \uppercase{Ljup\v{c}o Todorovski}\authorrefmark{3,2}}%
\address{Faculty of Mathematics and Physics, University of Ljubljana, Ljubljana, Slovenia}%
\address[2]{Department of Knowledge Technologies, Jo\v{z}ef Stefan Institute, Ljubljana, Slovenia}%
\address[3]{Faculty of Administration, University of Ljubljana, Ljubljana, Slovenia}

\markboth
{Lu\v{s}i\v{c} \headeretal: Meta-Model Framework for Surrogate-Based Parameter Estimation}
{Lu\v{s}i\v{c} \headeretal: Meta-Model Framework for Surrogate-Based Parameter Estimation}

\corresp{Corresponding author: jovan.tanevski@ijs.si}

\begin{abstract}%
The central task in modeling complex dynamical systems is parameter estimation. This task is an optimization task that involves numerous evaluations of a computationally expensive objective function. Surrogate-based optimization introduces a computationally efficient predictive model that approximates the value of the objective function. The standard approach involves learning a surrogate from training examples that correspond to past evaluations of the objective function. Current surrogate-based optimization methods use static, predefined substitution strategies to decide when to use the surrogate and when the true objective. We introduce a meta-model framework where the substitution strategy is dynamically adapted to the solution space of the given optimization problem. The meta model encapsulates the objective function, the surrogate model and the model of the substitution strategy, as well as components for learning them. The framework can be seamlessly coupled with an arbitrary optimization algorithm without any modification: It replaces the objective function and autonomously decides how to evaluate a given candidate solution. We test the utility of the framework on three tasks of estimating parameters of real-world models of dynamical systems. The results show that the meta model significantly improves the efficiency of optimization, reducing the total number of evaluations of the objective function up to an average of 77\%.
\end{abstract}


\titlepgskip=-15pt
\maketitle

\section{Introduction}\label{intro}

\PARstart{E}{stimating} the values of parameters of mathematical models of dynamical systems is often formulated as an optimization task with a computationally expensive objective function \cite{jaqamandanuser2006}. Given measurements of the behavior of a dynamical system, the task is to find values of model parameters that lead to a model simulation that closely fits the measurements. Computationally expensive simulation of the model is needed to assess the discrepancy between simulated and measured behavior of the observed system. Therefore, optimization approaches to parameter estimation can highly benefit from the use of surrogate-based optimization, which uses efficient approximation of the objective function. Such use of surrogates can thus substantially improve the efficiency of mathematical modeling.

Surrogate-based optimization solves optimization problems in situations where the resources for evaluating the objective function are limited. In computational domains, the limiting resource is most commonly computation time, which becomes critical when dealing with computationally expensive objective functions. The fundamental idea of surrogate-based optimization is to replace the computationally expensive objective function with a surrogate, i.e., a computationally efficient model that approximates the value of the true objective function. Surrogate-based methods employ machine learning algorithms for learning the surrogate model from training instances derived from the available evaluations of the true objective function.

Surrogate-based optimization can be deployed in two different application contexts. The first assumes a very limited number of evaluations of the true objective function. The aim of the surrogate model is to guide the selection of the most promising candidate solutions for evaluation. The Bayesian optimization approach \cite{jones1998} uses the surrogate model predictions and the corresponding confidences for selecting the next candidate solution that will be evaluated with the true objective function. The computational complexity of the selection process increases proportionally with the cube of the number of previously evaluated candidate solutions.

The use of Bayesian optimization is thus prohibitive in the second application context that assumes a large number of evaluations of the true objective function. The aim of the surrogate model, in this context, is to improve the efficiency of the optimization by replacing a large portion of the evaluations of the true objective function with evaluations of its surrogate. The parameter estimation task, addressed in this paper, fits this application context. The key component of the approaches applicable in this context is the substitution strategy that, for a given candidate solution, decides whether to evaluate it with the surrogate function or the true objective function \cite{jin2011}. Current approaches focus on maximizing the predictive performance of the surrogate model and use fixed, hard-coded substitution strategies \cite{bagherietal2015,dasetal2016,rammohan2015}.

In this paper, we design a meta-model framework for surrogate-based optimization with a substitution strategy that dynamically adapts to the space of evaluated candidate solutions. It includes two learning components: a component for learning a surrogate model of the true objective function and a component for learning a model of the substitution strategy. Additionally, the meta-model framework encapsulates the objective function, the surrogate model, the model of the substitution strategy, the history of evaluations and the learning components in a single, yet modular entity. An important consequence of the encapsulation is that the meta-model can be seamlessly coupled with an arbitrary optimization algorithm without any modification of the algorithm or the meta model. The latter replaces the true objective function and autonomously decides what function or model to use for the evaluation of a given candidate solution.

In our previous study \cite{luksic2017}, we show that learning the substitution strategy improves the overall performance of surrogate-based optimization. By learning the substitution strategy, instead of using a predefined one, the meta model is capable of solving complex numerical optimization problems while significantly reducing the number of evaluations of the true objective function. In this paper, we focus on the configuration of the learning components of the meta model. In particular, we conjecture that the selection of appropriate learning algorithms for the surrogate and substitution-strategy models significantly impacts the overall performance of surrogate-based optimization.

To test the validity of the conjecture, we perform an extensive empirical evaluation of different instances of the meta-model framework. Each corresponds to a pair of learning algorithms for training the surrogate, on one hand, and the substitution-strategy model, on the other. We select each algorithm in the pair among six alternative algorithms for learning predictive models, previously used in the literature on surrogate-based optimization---linear regression, decision trees, nearest neighbors, support vector machines, Gaussian processes and random forests---leading to 36 meta-model instances. In the first series of experiments, performed on synthetic benchmarks \cite{hansen2016}, we tune the parameters of each meta-model instance. In turn, we select the most successful instances that significantly outperform the others. The selected meta-model variants are evaluated in a second series of experiments on three real-world tasks of estimating the parameters of models of dynamical systems described by systems of coupled ordinary differential equations \cite{jaqamandanuser2006,kirk2016}.

We first describe in more detail the task of numerical optimization and parameter estimation. Next, we introduce and formally define the meta model, its components and parameters. We then lay out the setup for the empirical evaluation and report the results of our analyses. Finally, we provide a summary of our conclusions and outline directions for further research.

\section{Numerical optimization and parameter estimation}

We consider the task of numerical optimization involving a single, nonlinear objective function $\Fun$ in an unconstrained, continuous space $\fSupp$. The task is to find a solution $x^\ast \in \fSupp$ that leads to the extremum of the objective function $\Fun: \fSupp \to \R$. The objective function can be either minimized or maximized: in the former case, the result of optimization is $x^\ast = \argmin_{x\in\fSupp} \Fun(x)$.

If the analytical solution for the minimum of $\Fun$ is intractable, numerical methods are applied. These methods can belong to two groups: local and global optimization methods. While local methods \cite{nocedal2006} are efficient, they suffer from myopia, i.e., the tendency to end up in a local extreme point in the neighborhood of the initial point. On the other hand, global methods \cite{pinter1995} are concerned with finding the global optimum point and use different strategies for sampling the solution space. To improve their efficiency, they are often coupled with surrogates.

Parameter estimation aims at finding values of the parameters of a given model of a dynamical system that result in a model simulation that closely fits a given set of measurements of the observed system behavior. Models of dynamical systems are usually formalized as systems of coupled ordinary differential equations $ \dot{y} = G(y, x)$ \cite{murray1993}, where $ y $ denotes the vector of the observed state variables of the dynamical system, $ \dot{\boldsymbol{y}} $ is the vector of the time derivatives of the state variables, $ G $ is the function representing the model structure, and $ x $ denotes the vector of the real-valued constant parameters of the model. Given an initial condition $y_{t_0}$, i.e., the value of $ y $ at the initial time point $t_0$, the model simulation leads to a set of trajectories of the dynamical change of the state variables $ y $ through time. Analytical solutions of systems of coupled ordinary differential equations describing complex real-world models are rarely an option, so computationally expensive numerical approximation methods for integration (simulation) are typically applied.

\begin{figure*}[!ht]
\centering
\includegraphics[width = \xfigwd]{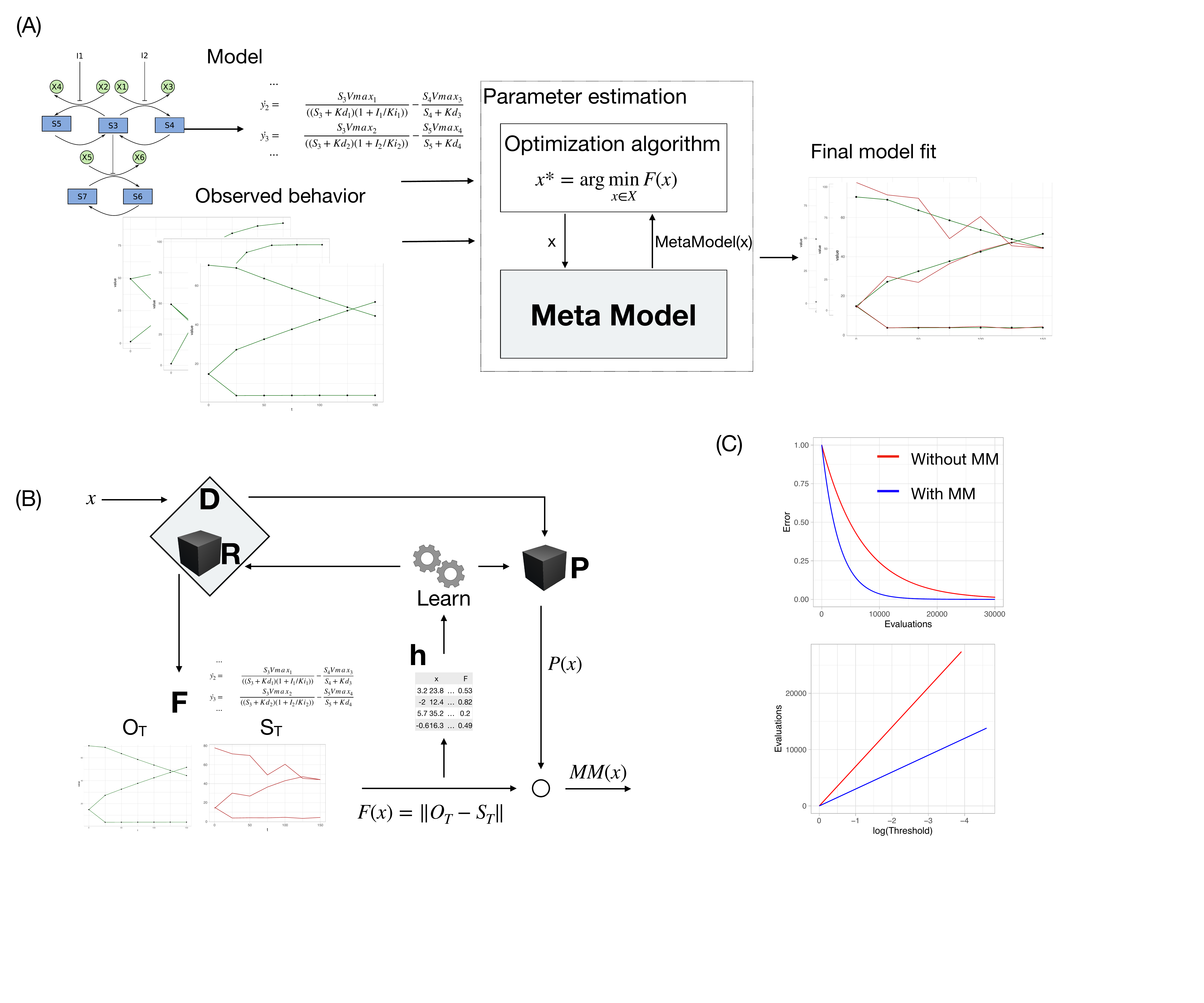}
\vspace*{-2cm}
\caption{The meta-model framework for surrogate-based parameter estimation in dynamical systems.
(A) The parameter estimation task takes at input (1) the model of the biological dynamical system in its equation-based formulation (top-left boxes) and (2) the measured trajectories of the observed system behavior. It uses an optimization algorithm, coupled with the meta model for surrogate-based optimization, to find an optimal value of parameters $ x^\ast $ that leads to a final model, the simulation of which closely fits the observed system behavior.
(B) The meta model takes the candidate solution $ x $ at input. The decision function D uses the relevator $ R $ to evaluate the relevance of $ x $ and choose what function to use for its evaluation. For highly relevant $ x $, it will opt for the true objective function $ F $, that involves simulation of the model with parameters $ x $ and comparison of the simulated behavior $ S_T $ with the observations $ O_T $. The $ x $ and $ F(x) $ will be stored in the history part of the meta model $ h $. This part contains the evaluations of the objective functions and is used as a training set for the surrogate model $ P $ and the relevator $ R $. A candidate solution $ x $ with low relevance will be evaluated using the surrogate $ P $.
(C) The utility of the meta model is assessed by comparing the two convergence curves (upper graph) of the plain optimization algorithm (without MM) with the surrogate-based optimization using the meta model (with MM). The convergence curve depicts the optimal value of the objective function (y axis) as the number of function evaluation (x axis) increases: the lower the convergence curve, the better is the utility. The transposed convergence curve (lower graph) depicts the number of evaluations (y axis) necessary to achieve a certain threshold value of the objective function (x axis, logarithmic scale). Lower curve indicates better utility.}
\label{figure1}
\end{figure*}

The task of parameter estimation (Figure~\ref{figure1} (A)) can be formalized as follows. Given the measured behavior $O_T$ of the system variables at time points $T$, the task is to maximize the likelihood of the observed behavior given a particular value of $ x $, i.e., $F(x) = -\mathcal{L}(O_T|x)$, where $\mathcal{L}$ is a likelihood function. In practice, due to the complexity of the models considered, the likelihood-based function is approximated by a least-squares function $F(x)= \lVert O_T-S_T \rVert$, where $S_T$ denote the simulated behavior of the system variables at time points $T$. Recall however, that $S_T$ is obtained by using a computationally intensive method for integrating differential equations, often leading to inefficient optimization and poor optima.

This is especially true in the process of discovery of knowledge about the complex behavior and function of biological systems. This often involves mathematical modeling of dynamical systems from observational data \cite{kirk2016}, with a key aspect being the task of parameter estimation \cite{jaqamandanuser2006}. Regarding the choice of a parameter estimation method for problems coming from the domain of systems biology, global stochastic and hybrid methods based on metaheuristics are considered most promising in the literature \cite{ashyraliyev2009,chou2009}. These methods require a large number of objective function evaluations, which makes them ideal candidates for applying surrogate-based optimization. 

\section{Meta-model framework for surrogate-based optimization}

We first introduce the mathematical \emph{meta-model} framework for surrogate-based optimization in its abstract form. This part is accompanied by a graphical overview of the framework depicted in Figure~\ref{figure1}(B). We then gradually proceed by specifying the machine learning components of the framework. Finally, we introduce a relevance-based surrogate management strategy that allows the meta model to make autonomous decisions about which function or model to use for evaluating a given candidate solution.

\subsection{Meta-model framework}

Our meta-model framework consists of four components. The first is the objective function $\Fun$ that is the subject of optimization. The second component is the surrogate model $\Surr: \fullSupp \to \R$ that for a given candidate solution (or a query point) $x$ and based on the history of previous meta-model evaluations $\hist$ computes $ S(x,\hist) $, the surrogate approximation of $\Fun(x)$. The third component is the decision function $\Dec: \fullSupp \to \{0,1\}$ that implements the dynamic substitution strategy: for a given $x$ and based on $h$, it decides whether to use the objective function $\Fun$ (output 1) or its surrogate $\Surr$ (output 0). Finally, the fourth important component of the meta-model framework $\hist$ is the ``\emph{history of evaluations}'', where the data on past meta-model evaluations is kept. The evaluation history is a finite sequence of vectors: $\hist \in \hSupp$ where $\dimension$ is the dimension of the optimization problem and $\hDim$ the dimension of additional data being kept in the history.

More formally, the function $\MM: \fSupp \to \R$ is defined by three functions and a history of evaluations $(\Fun,\Surr,\Dec;\hist)$, corresponding to the components of the \emph{meta model}: 
	\begin{itemize}
		\item objective function $\Fun: \fSupp \to \R$,
		\item surrogate function $\Surr: \fullSupp \to \R$,
		\item decision function $\Dec: \fullSupp \to \{0,1\}$.
		\item evaluation history $\hist: \hSupp$.
	\end{itemize}

Given this components of the meta-model framework, the function $\MM$ is defined as\footnote{Note that the alternative notation $\MM(x;\hist)$ will be used whenever we want to emphasize the modification of the evaluation history.}
\begin{equation}
	\MM(x)=
	\begin{cases}
	F(x);&\Dec(x, \hist)=1\\
	S(x,\hist);&\Dec(x, \hist)=0
	\end{cases}
\label{eq-mm}
\end{equation}

While the functions $\Fun,\Surr,\Dec$ can be arbitrary black boxes, we assume that the evaluation history of the meta model is updated after every evaluation. For our current needs, the evaluation history records the query point $x=(x_1, \ldots, x_k) $, the result of the meta model $ \MM(x) $ and the value of the decision function $ \Dec(x,\hist) $. If we denote the $r$-th evaluation of the meta model at the point $x^{(r)}$with $\MM(x^{(r)};\hist_r)$ and the next one with $\MM(x^{(r+1)};\hist_{r+1})$, then $\hist_{r+1}$ is the extension of $\hist_r$ with the vector $(x_1^{(r)}, \ldots, x_k^{(r)}, \MM(x^{(r)}; \hist_r), \Dec(x^{(r)}, \hist_r))$. The last two values are used to define the values of the targets in the data sets for learning the predictive models in the components $\Surr$ and $\Dec$.

\begin{algorithm}[t]
\caption{The meta model for surrogate-based optimization. At input it takes the candidate solution $x$ and at output it returns its evaluation $\MM(x)$.}
\label{alg-mm}
\begin{algorithmic}
\STATE{Input $ x $: candidate solution to be evaluated; $ x \in \fSupp $}
\STATE{Output $ {\it value} $: evaluation of the candidate solution $ x $}
\STATE{Parameters $ T_1 $ and $ T_2 $ control the number of training examples $ T_1 k + T_2 $}
\STATE{Parameters $ I_1 $ and $ I_2 $ control the minimum number of new examples $ I_1 k + I_2 $}
\STATE{Parameter $ r $ controls the desired surrogate replacement rate}
\STATE
\IF {$ ({\it nEvalsF} - (\dimension \cdot T_1 + T_2)) \, \% \, (\dimension \cdot I_1 + I_2) = 0 $}
	\STATE {$ S.{\rm learn}(h) $}
	\STATE {$ R.{\rm learn}(h) $}
\ENDIF
\IF {$ {\it nEvalsF} < \dimension \cdot T_1 + T_2 $}
    \STATE {$ {\it decision = 1} $}
\ELSE
    \STATE {$ {\it threshold}.{\rm update}(h,r) $}
    \STATE {$ {\it decision} = R.{\rm evaluate}(x) > {\it threshold} $}
\ENDIF
\IF {$ {\it decision} = 1 $}
	\STATE {$ {\it value} = F.{\rm evaluate}(x) $}
	\STATE {$ {\it nEvalsF} = {\it nEvalsF} + 1 $}
\ELSE
	\STATE {$ {\it value} = S.{\rm evaluate}(x) $}
\ENDIF
\STATE {$ h.{\rm add}(x, {\it value}, {\it decision}) $}
\STATE {$ {\it nEvals} = {\it nEvals} + 1 $}
\RETURN {\it value}
\end{algorithmic}
\end{algorithm}

Algorithm~\ref{alg-mm} presents the meta model for evaluating a given candidate solution $x$. First, the meta model checks whether there are enough evaluations of the true objective function (variable {\it nEvalsF}) in the evaluation history $ \hist $ for training (or re-training) the surrogate and relevator models (functions {$ S.{\rm learn}(h) $} and {$ R.{\rm learn}(h) $}, see subsection~\ref{relevator}). To this end, parameters $ T_1 $, $ T_2 $, $ I_1 $ and $ I_2 $ are being used; we provide further explanation in subsection~\ref{surrogate}. Next, if the predictive models are not available yet, the meta model opts for evaluating the true objective function (the variable {\it decision} is set to have a value of 1). Otherwise, based on the evaluation history $\hist$ and the parameter $ r $, it updates the decision threshold (see subsection~\ref{relevator}) and uses it to decide whether to use the surrogate model (equivalent to $S.{\rm evaluate}$ in the pseudo code) or the true objective function $\Fun$ to evaluate $x$. It stores the evaluation in the variable {\it value}, which is the result of the meta model function. The meta model updates the counter {\it nEvalsF} that keeps track of the number of evaluations of the true objective function.

\subsection{Surrogate}\label{surrogate}

The surrogate function $\Surr$ takes care of learning, updating and evaluating the surrogate predictive model $ \Predictor: \fSupp \to \R $. In the rest of the paper, we use $\Surr$ when talking about the whole and $\Predictor$ when talking about the predictive model component of $\Surr$. There are three important aspects to be considered when constructing a good surrogate function: the type of the predictive model, the size of the training set used for its initialization and the frequency of the model updates. 

We aim at selecting a surrogate predictive model that closely approximates the objective function and can be evaluated efficiently. Because we do not wish to additionally sample data for $\Fun$, we only rely upon the data stored in the evaluation history $\hist$. This is particularly beneficial when working with population-based methods, as the samples are more concentrated in the current area of our population, where we want better accuracy of our prediction model. Moreover, the efficiency of the surrogate function depends upon the trade-off between the frequency of surrogate learning and the size of the training set. Having a high update frequency is desirable since the surrogate then always takes into account the most recent history of evaluations. On the other hand, frequent surrogate updates are unprofitable unless the learning time is fairly low compared to the evaluation time of the true objective function. To this end, we introduce user-defined parameters that determine the number of true object evaluations between the consecutive surrogate updates (parameters $ I_1 $ and $ I_2 $ in Algorithm~\ref{alg-mm}).

A larger training set substantially slows down the algorithm by increasing the time needed to learn the surrogate model, which in turn negatively impacts the time performance of the meta-model. We attempt to solve the problem with \emph{filtration} of the history of evaluations. In order to minimize the error of our prediction model, only evaluations of $\Fun$ are used, which are easily extracted from the evaluation history because we additionally keep the values of $\Dec$ (records from the evaluation history are extracted, where $\Dec(x,\hist)=1$). Furthermore, when using the meta model in conjunction with a population-based method, the population slowly converges towards the minimum of the true objective function. We want the surrogate to be more precise in the current area of the search, which coincides with the most recent evaluation points. By focusing only on the most recent evaluations, we additionally emphasize points with a lower value of the objective, meaning less noise from the high value outliers, which are less relevant for learning the surrogate, since they correspond to non-optimal points. Therefore, in our implementation, the training set includes a user-defined number of the most recent points from the history of evaluations (parameters $T_1$ and $T_2$ in the Algorithm~\ref{alg-mm}). Additional filtration schemes are possible, for instance considering only the points with the lowest values of the objective function or a combination of the two. While all schemes are subsumed by adding a weight function, that introduces a large amount of additional parameters.

\subsection{Relevator}\label{relevator}

Selecting a suitable decision function is of vital importance. As shown in our previous work\cite{luksic2017} a simple \emph{uninformed} decision function, which uses the step number as its only argument, performs poorly on harder optimization problems. One possible way of constructing a more successful decision function is by predicting how relevant the point will be for our optimization algorithm. It is based on the idea that points of high ``\emph{relevance}'' for the optimization algorithm need to be predicted more accurately in order not to slow down progress. Thus, the evaluation of the most relevant points should be performed using the true objective function while less relevant points can be evaluated using the surrogate model.

Taking a decision based on the relevance of a point, brings up two issues. First, how do we formally define point relevance and second, how can the relevance of a point be estimated before evaluating the objective function. The point relevance can be calculated with a function passed to the meta model as an argument. However, in our current implementation, we define the relevance of a point $x\in\fSupp$ relative to the lowest value of the objective function seen so far: the closer is the value to the lowest value, the higher is its relevance. In particular, if we use $f = (f_1\cdots,f_m)$ to denote the vector of values of $\Fun$ in the evaluation history, we define the point relevance as 
\begin{equation}
    \relevance(x,f) = 
    \begin{cases}
        \left(1+\frac{\Fun(x)-\min_i f_i}{\avg_i f_i-\min_i f_i}\right)^{-1}&F(x)\geq\min_i f_i\\
        1&F(x)<\min_i f_i
    \end{cases}
\end{equation}

\noindent
The relevance of a point is bound by definition in the interval $[0,1]$ where the value of $0$ corresponds to a point of low relevance and $1$ to a point of high relevance and points close the current average value get mapped close to $0.5$.

Having defined point relevance, we have to resolve the remaining issue of its estimation without evaluation of the objective function. We could use the same prediction model as $\Surr$ to approximate $\Fun$ and then calculate the relevance by simply replacing $\Fun(x)$ with $\Predictor(x)$ in the formula above. Instead, we have decided to learn a separate model for predicting the point relevance. This not only gives us a much wider array of possible meta models but also allows us to dynamically adapt our relevance prediction model to the optimization task at hand. We refer to this model as the {\em relevator}.

To reduce the number of parameters in the framework, when constructing the training set for the relevator, we have decided to reuse the same filtering scheme as with the surrogate. Thus, to learn the relevance function, we construct the vector $f$ by using the evaluation history $\hist$. However, we decided to only include values of $\Fun$ present in our training set. Using the whole history of evaluations tends to increase the average of $f_i$ making it difficult to distinguish relevant points as their relevance moves closer towards $1$. This problem is reduced by only using ``recent values'' present in out filtered train set. 

As with the surrogate function $\Surr$, a call of the relevator $\Rel(x, \hist)$ not only predicts the value of $\relevance(x, f)$, where $f$ is taken from the filtered $\hist$, but also learns and updates the relevance prediction model whenever needed. In order to reduce the number of parameters we reuse the surrogate parameters for update frequency.

In addition to the relevator, the decision function includes a decision threshold that distinguishes the points with high relevance, which should be evaluated using the true objective, from points with low relevance, which should be evaluated with the surrogate. To allow for dynamic change of the threshold value $\threshold$, we define it as a function of the evaluation history $\hist$.

Thus, the decision function of a \emph{relevator} meta model is the indicator function $\textbf{1}[\Rel(x,\hist) > \threshold(\hist)]$, where $\Rel: \fullSupp \to [0,1]$ is the relevance estimate of point $x$ given the history of evaluations $\hist$ and $\threshold: \hSupp \to [0,1]$ is a dynamical relevance threshold function.
\begin{equation}
	\Dec(x)=
	\begin{cases}
	1; &\Rel(x,\hist) > \threshold(\hist) \\
	0;& \Rel(x,\hist) \leq \threshold(\hist)
	\end{cases}
\end{equation}

We implement the dynamical relevance threshold using an iterative update procedure with the goal to control and locally adjust the frequency of surrogate evaluations. By considering the user-defined number of most recent evaluations, we can either raise or lower the threshold after every meta model evaluation in order to increase or decrease the frequency of surrogate evaluations to (locally) achieve the desired substitution rate.

\section{Results}

In this section, we present the setup and the results of two series of experiments with the proposed meta-model framework. In the first series, we use 45 standard benchmark problems for numerical optimization to tune the parameters and evaluate the performance of the framework with different meta-model instances. Based on the comparison of their performance, we identify the machine learning methods that lead to suitable surrogate and substitution-strategy (relevator) models. The most successful among them are evaluated on a second series of experiments on three tasks of estimating the parameters of three real-world models of dynamical systems from the domain of systems biology.

\subsection{Meta-model tuning and selection}

The construction of both the surrogate and the relevator functions for the meta model can be readily framed as a regression problem. We considered combinations of six different methods for learning: linear regression (LINEAR), regression tree with variance reduction and reduced-error pruning (TREE), k-nearest neighbors with k=5 (KNN), Gaussian processes with squared exponential covariance (GP), support vector machines, $\epsilon$-SVM with RBF kernel (SVM), and Random Forest with 100 trees (RF). We used the default parameters from the Weka implementation for each method \cite{witten2016}. The selection of the machine learning methods is based on the state-of-the-art studies of surrogate-based optimization approaches \cite{bagherietal2015,dasetal2016,rammohan2015} that uses mostly Gaussian processes, followed by Random Forest and support vector machines. To check the utility of other, more efficient machine learning methods, we decided to include also linear regression, nearest neighbors and regression trees.

\begin{table}[!h]
\caption{Values considered for tuning the five parameters of the meta model.}
\centering
\begin{tabular}{r|l}
\hline
Parameter & Candidate values \\
\hline
$T_{1}$ & 0, 10, 25, 50  \\
$T_{2}$ & 100, 200, 500 \\
$I_{1}$ & 0, 10, 25, 50  \\
$I_{2}$ & 50, 100, 200 \\
$r$ & 0.3, 0.4, 0.5, 0.6, 0.7, 0.8 \\
\hline
\end{tabular}
\label{tabGrid}
\end{table}

For each of the 36 surrogate-relevator combinations, we tuned the five parameters of the meta model by using COCO, the platform for comparing numerical optimization methods in a black-box setting \cite{hansen2016}. The parameters were tuned by using grid search with values as shown in Table~\ref{tabGrid}. The parameters $T_1$ and $T_2$ are used to calculate the training set size and the parameters $I_1$ and $I_2$ are used to calculate the interval for rebuilding the surrogate as $\dimension \cdot T_1 + T_2$ and $\dimension \cdot I_1 + I_2$. Both sizes are relative to the dimension of the problem $\dimension$. The last parameter is $r$, the desired rate of substitution of the objective function with the surrogate model.

The widely-used COCO platform contains a set of black-box optimization functions that are used as benchmarks problems for numerical optimization. From this set, we selected 15 functions from three different classes of problems that resemble real-world parameter estimation tasks: uni-modal functions with high conditioning, multi-modal functions with adequate global structure and multi-modal functions with weak global structure. Within each class the functions differ by levels of deceptiveness, ill-conditioning, regularity, separability and symmetry. Each optimization function can be generalized to a different number of dimensions. We consider instances of each function in 5, 10 and 20 dimensions. Thus, we use a total of 45 benchmark optimization functions. The best set of parameters for each surrogate-relevator pair is selected that maximizes the improvement in performance $\pi$ relative to a baseline optimization method without surrogates. 

Regarding the choice of a parameter estimation method, global stochastic and hybrid methods based on metaheuristics are considered as most promising in the literature \cite{ashyraliyev2009,chou2009}. Out of the many different methaheuristic methods, Evolutionary Strategies and Differential Evolution have been identified as the most successful \cite{sun2012,tashkova2011} in the intended domain of application in this work -- systems biology. We use Differential Evolution (DE) \cite{stornandprice1997} as the baseline optimization method. It is a staple evolutionary and population-based method that has consistently shown robust and reliable performance on various problems in many domains \cite{chakraborty2008}.

For each of the 45 functions we establish the baseline performance of DE without meta model by running it with a budget of $1000 \cdot \dimension$ evaluations. The performance improvement of a meta-model on a single benchmark function $f$ is then calculated as $\pi_f = \max(0, 1 - M_f/N_f)$. Note that $N_f$ represents the number of evaluations needed to reach the minimum value $V_{f}$ without surrogates, while $M_f$ is the number of true function evaluations needed to reach the same minimum of $V_{f}$ using the meta model. The overall performance improvement of a meta-model is then calculated as $\pi = \bar{\pi_f} + s$, where $\bar{\pi_f}$ is the average performance improvement on all 45 functions and $s$ is the proportion of all functions $ f $ with $\pi_f > 0$.

\begin{figure}[!t]
(A)\\
\includegraphics[width=\linewidth]{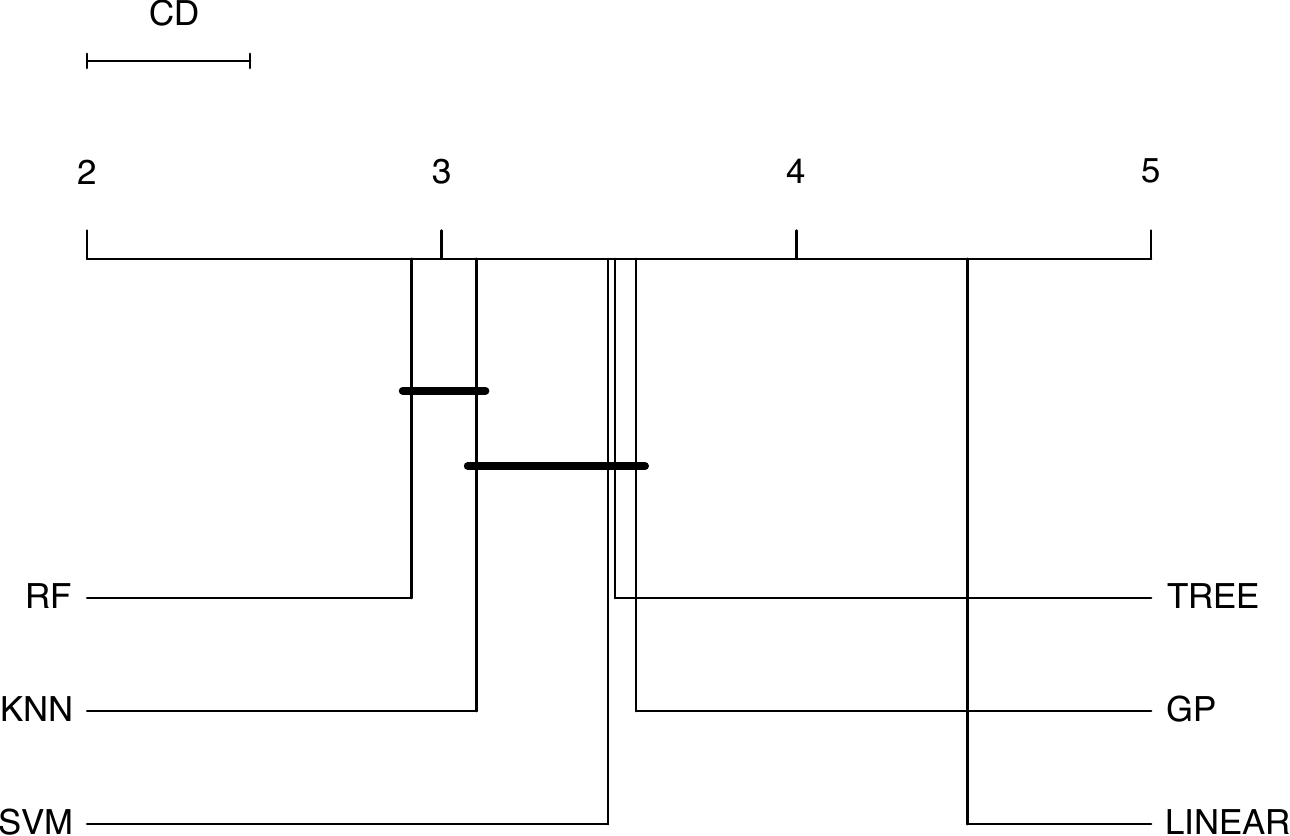}\\
(B)\\
\includegraphics[width=\linewidth]{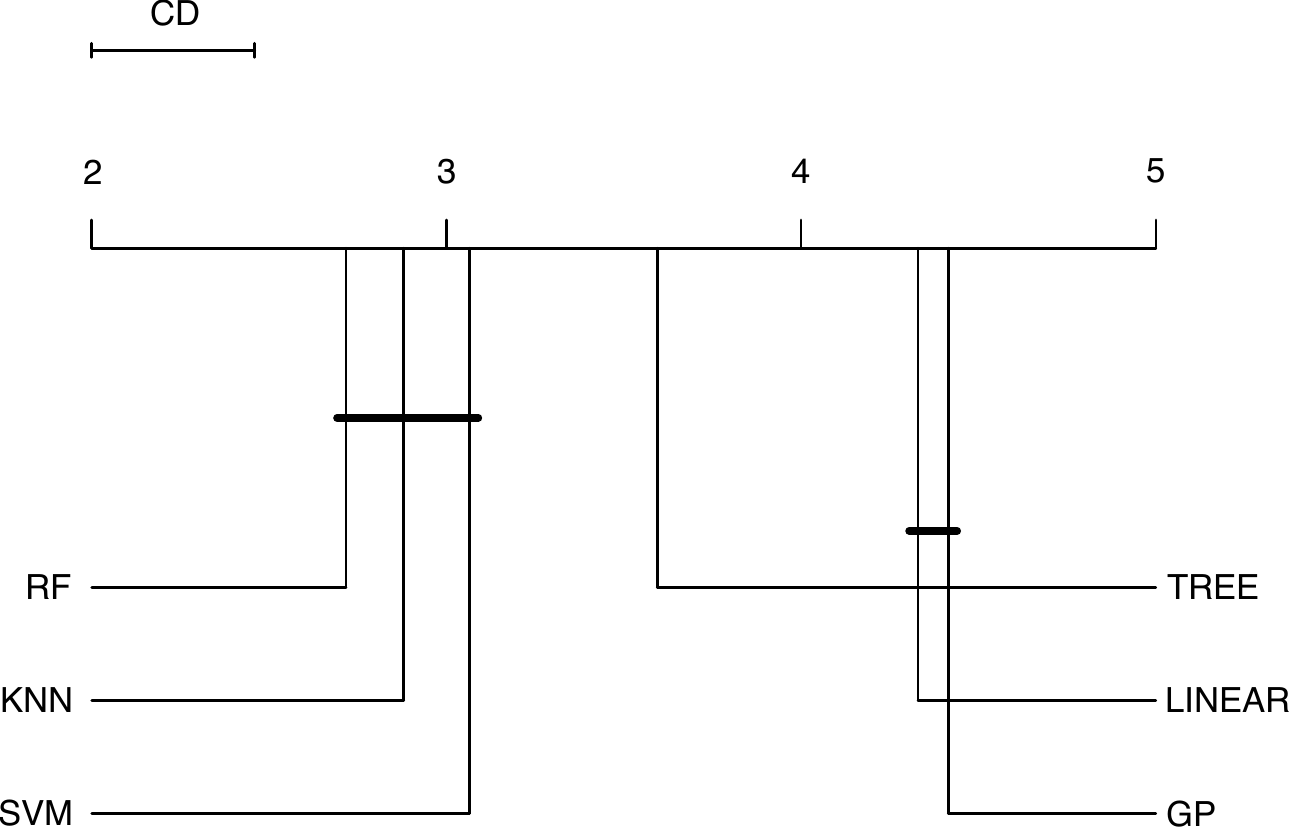}
\caption{Average ranks of the meta-model instances on the 45 numerical optimization benchmarks. The meta-model instances are grouped according to the algorithm used to learn the surrogate (A) and the substitute-strategy model (B). Each point in the diagram corresponds to the ranks of the meta-model instances in the group averaged over all the 45 benchmarks and 6 algorithms used for the second learning component.}
\label{figARD}
\end{figure}

For each surrogate-relevator pair we selected the parameters that resulted in the best performance improvement. While we cannot draw any firm conclusions about the best size of the training set and the rebuild interval, the results show the potential of the meta model to achieve a very high replacement rate of evaluations of the true objective function with evaluations of the surrogate function. For parameters $T_1$ and $I_1$, all possible values were selected an equal number of times as the best configuration. The lowest values were most frequently selected for $T_2$ and $I_2$ (17/36 and 14/36). Most importantly, the highest possible value of 0.8 or the parameter $r$ was selected in 24 out of 36 best configurations.

\begin{figure*}[!ht]
\centering
\includegraphics[width=.9\textwidth]{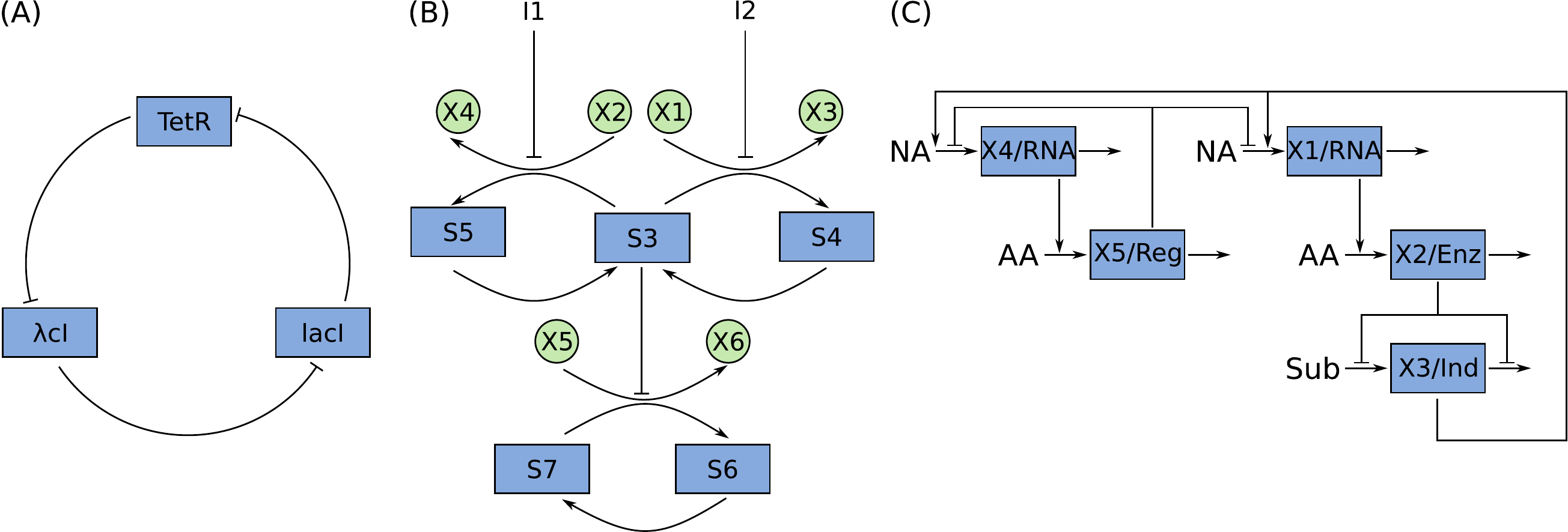}
\caption{Diagrams of the three models of dynamical biological systems. (A) Synthetic oscillatory network - repressilator; (B) Metabolic NAND gate; and (C) S-system model of a genetic network. The rectangles represent observed and modeled variables. The arcs ending with an arrow ($\rightarrow$) represent interactions with positive regulation while the arcs ending with a bar ($\dashv$) represent interactions with negative regulation.}
\label{fig-pr}
\end{figure*}

To analyze the impact of the selection of the algorithms for learning the surrogate and the relevator model, we performed Friedman's rank sum test and a Nemenyi post-hoc analysis \cite{demsar2006}. To perform the tests, we grouped the methods in two ways. First, we grouped the meta-model instances in 6 groups according to the algorithm used for learning the surrogate. The second grouping is based on the algorithm used for learning the substitution-strategy model. For both groupings, the Friedman test checks the validaty of the null hypothesis that all the groups of meta-model instances perform equally well. In both cases, the null hypotheses is rejected with a p-value of less than $ 2 \cdot 10^{-16} $, leading to a conclusion that the selection of the algorithm for learning the surrogate or the relevator has a significant impact on the performance of the meta model. Furthermore, the post-hoc analysis uses the Nemenyi test to investigate the significance of the existing differences by calculating the critical distance between the average ranks of the groups. The performances of the two groups differ significantly, if their difference is larger or equal to the critical distance. In both groupings, the critical distance at the significance level of 0.05 equals 0.4594.

Figure~\ref{figARD} employs average-rank diagrams to summarize the results of the comparative analysis. The horizontal axis of the average-rank diagram corresponds to the rank of the group of meta-model instances: the top-ranked group is on the left-most position on the axis. The line above the axis, labeled CD, depicts the value of the critical distance. The groups of meta-model instances with utilities that are not significantly different are connected with thick lines below the axis.

The average-rank diagram in Figure~\ref{figARD}(A) groups the meta-model instances according to the algorithm used to learn the surrogate. Considering the significance of the pair-wise differences between the groups, we can exclude linear regression as an algorithm leading to meta-model instances with significantly inferior performance. The average-rank diagram in Figure~\ref{figARD}(B) groups the meta-model instances with respect to the algorithm for learning the substitution-strategy model. Meta-model instances using Random Forest, nearest neighbors and support vector machines for learning the relevator significantly outperform the other meta-model instances using the other three algorithms. The comparison of the two graphs reveals that the meta model is more sensitive to the selection of the algorithm for learning the relevator then the selection of the algorithm for learning the surrogate. This is an important insight showing that the choice of the model for the dynamical substitution strategy adapted to the problem at hand has a significant impact on the utility of surrogate-based optimization.

\subsection{Estimating the parameters of dynamical biological systems}

We are interested in the performance of the meta-model in the case of the real-world problem of estimating the parameters of models of dynamical biological systems. We selected three dynamical biological systems with varying degrees of complexity (shown in Figure~\ref{fig-pr}). The three systems have been well studied in terms of their dynamical properties and identifiability \cite{buse2010,gennemarkwedelin2007}.

The first system is a synthetic oscillatory network of three protein-coding genes interacting in an inhibitory loop, known as the Repressilator, modeled by Ellowitz and Leibler \cite{elowitzleibler2000}. The system is modeled by six variables and four constant parameters. The time-series data for this problem is obtained by numerical integration of the system of ordinary differential equations with the parameter values reported in \cite{elowitzleibler2000} for 30 integer time-points. The objective function used is the sum of squared errors between the simulated trajectories of the model and the available data.

The second system is a metabolic pathway representing a biological NAND gate \cite{arkinross1995}. The model is represented by a set of five ODEs with 15 constant parameters. Observation data for the metabolic pathway model was obtained from \cite{gennemarkwedelin2007}. It consists of 12 sets of observations obtained by simulating the model using 12 different pairs of input step functions $(I1, I2)$ sampled uniformly at 7 time points. The objective function used is the negative log-likelihood of the simulated trajectories of the model and the observations, summed across all datasets.

\begin{figure*}[!ht]
(A) \hspace*{0.8cm} Relevator: SVM \hspace*{3.8cm} Relevator: RF \hspace*{3.5cm} Relevator: KNN \\
\includegraphics[width=\linewidth]{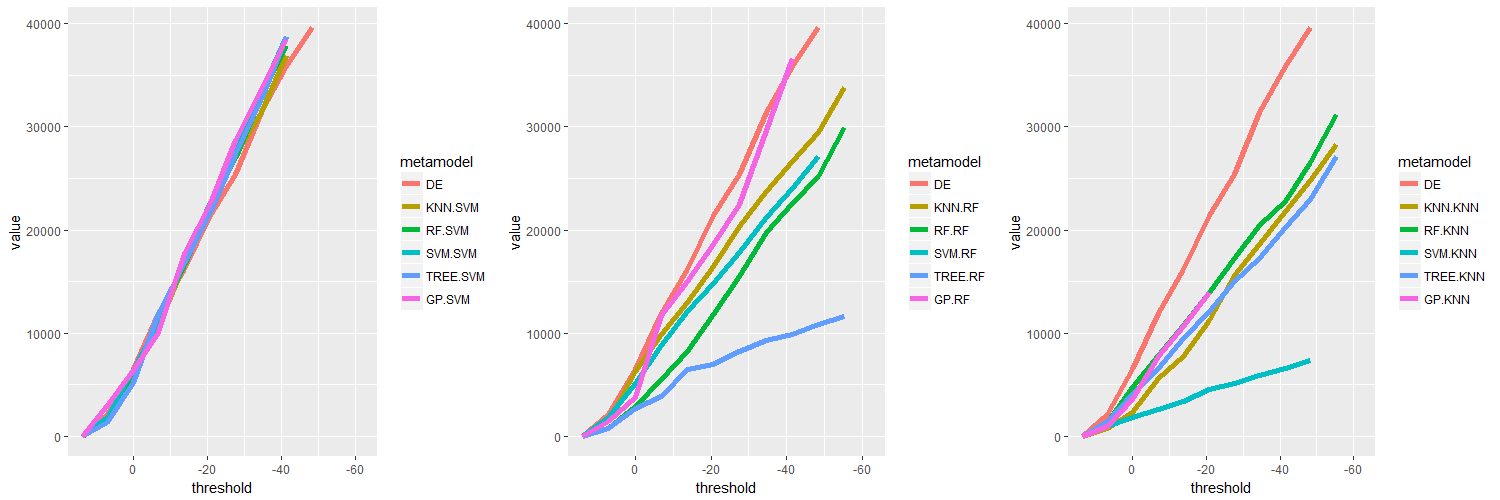}
(B) \\
\includegraphics[width=\linewidth]{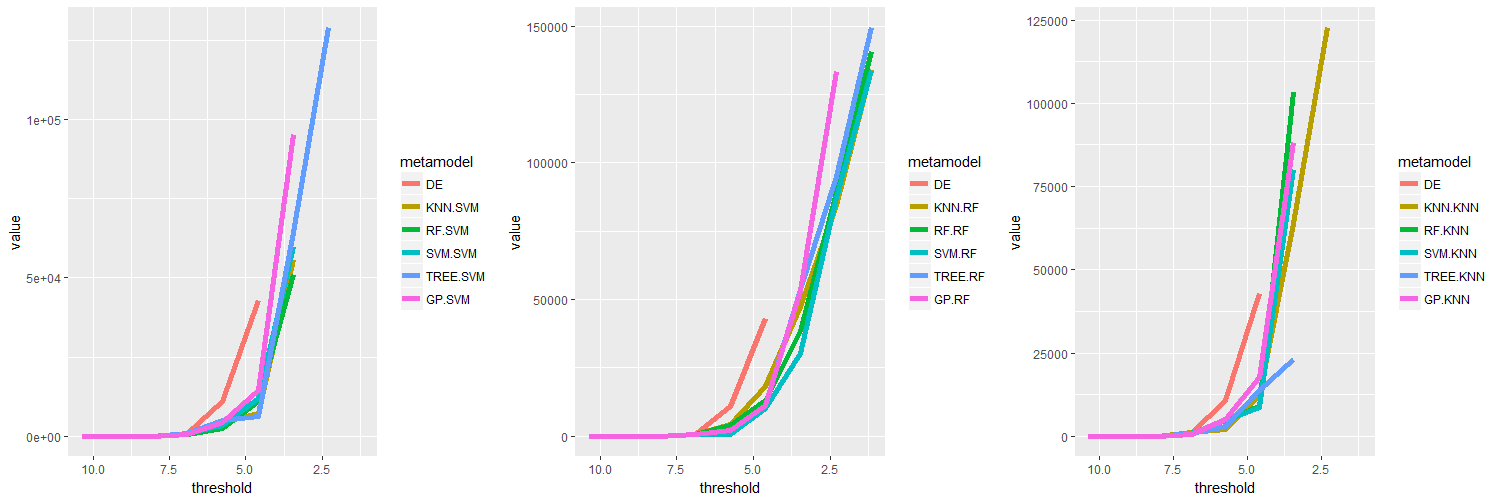}
(C) \\
\includegraphics[width=\linewidth]{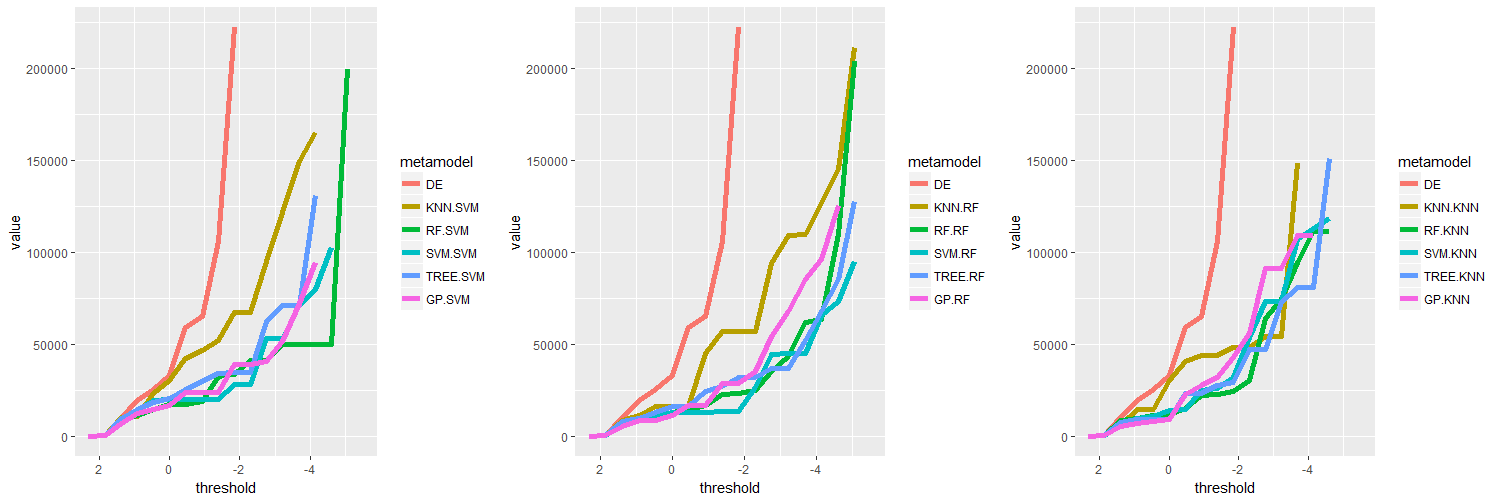}
\caption{Transposed convergence curves for the three parameter estimation problems: (A) Repressilator, (B) Metabolic pathway and (C) S-System model of genetic network. Optimization without a meta-model (DE) and a meta-model with different surrogate-relevator pairs (each column corresponds to one of the methods for learning the relevator). The curves show the number of true objective function evaluations needed to reach a certain objective value threshold. Points are missing from the end of some of the curves if that method did not reach the threshold in the allocated total number of evaluations.}
\label{figInv}
\end{figure*}

The third system is a genetic network \cite{kikuchi2003}. The system is represented as a five variable S-system model with 23 constant parameters. The observation data for the S-system model was obtained from \cite{gennemarkwedelin2007}. It consists of 10 sets of observations obtained by simulating the model using 10 different sets of initial conditions for all variables, sampled uniformly at 11 time points. The objective function is a log-transformation of the negative log-likelihood function, with preserved order and mapping $0$ to $0$. As in the previous system, the objective function was summed across all datasets.

To establish a baseline, we ran Differential Evolution (DE) without a meta-model on each problem with a budget of $10000 \cdot \dimension$ evaluations, where $\dimension$ is the dimensionality of the problem, i.e., the number of constant parameters in the system. For the meta-model we considered the 5 methods for learning a surrogate and the 3 methods for learning a relevator function, that were shown to have statistically significant performance improvements over the other methods. In all cases the parameter estimation was repeated 10 times with different random seeds.

Figure~\ref{figInv} shows the transposed convergence curves for the optimization runs (minimum of 10 restarts) without using a meta-model (DE) and using a meta-model with different surrogate-relevator pairs. Recall that the transposed convergence curve depicts the number of evaluations (y axis) necessary to achieve a certain value of the objective function (x axis, logarithmic scale) and thus, lower curve indicates faster convergence of the method. In all the graphs, the red curve corresponding to the optimization without the meta model is above the others indicating the superior convergence of the methods using the meta model. The exception to this rule are the meta models with the SVM relevator (depicted in the top left graph of the Figure~\ref{figInv}) that fail to outperform the baseline DE method. The graphs in the second column show the consistent superiority of the meta models using RF method as a relevator: for all the tasks, these methods achieve the lowest values of the objective function. For the repressilator task, the meta models using surrogates based on decision trees lead to fastest convergence; for the other meta models with tree-based and RF repressilators have superior convergence. Note finally that the meta models using the SVM relevator also consistently outperform the baseline, but the curves are shorter when compared to the meta models using RF relevator, indicating inferior utility with respect to the obtained value of the objective function.

More importantly, the meta-model achieves a significant improvement in the speed of convergence. To compare the convergence behavior of the meta-model to the baseline (DE) across all problems, we performed Page's trend test for ordered alternatives as proposed by Derrac et al. \cite{derrac2014} on 20 uniformly distributed cut points along the log-transformed convergence curves. We test the null hypothesis that the difference between two curves (minimum across 10 restarts) does not increase with the number of evaluations, i.e. there is no difference in the speed of convergence. Table~\ref{tabPage} shows the p-values obtained for the different surrogate-relevator pairs; note that the values below $0.01$ are emphasized in bold letters.

\begin{table}
\caption{p-values for the statistical significance of the difference between the convergence curve of a meta-model instance and the convergence curve of the no-surrogate method.}
\centering
\begin{tabular}{l|r|r|r|r|r}
\hline
$\frac{\text{\Surr} \rightarrow}{\downarrow \text{\Rel}}$ & TREE & KNN & GP & SVM & RF  \\
\hline
KNN & \textbf{3.69e-3} & \textbf{3.04e-5} & 0.504 & 0.372 & \textbf{5.87e-6} \\
SVM & 0.399 & 0.437 & 0.644 & 0.528 & 0.704 \\
RF & \textbf{5.82e-6} & \textbf{4.09e-13} & \textbf{4.98e-3} & \textbf{1.26e-8} & \textbf{4.26e-9} \\
\end{tabular}
\label{tabPage}
\end{table}

Page's trend test indicates that in terms of improvement in the speed of convergence, Random Forest is the superior choice for the relevator of a meta-model when combined with any choice of method for learning the surrogate function. The use of the k-nearest neighbors method for the relevator also results in more efficient convergence, however only when combined with certain methods (i.e., TREE, KNN and RF) for learning the surrogate function.

Both Random Forest and k-nearest neighbors are conservative estimators in that they are limited in their ability to extrapolate predictions for candidate solutions with feature values outside the space covered by solutions in the training set. We conjecture that this property of the surrogates and the relevator is exploited by the optimization method to efficiently explore non-optimal parts of the objective space, which improves the convergence. This property also reduces the possibility of error from evaluating the surrogate function when exploring parts of the solution space that have high potential for optimality.

We further analyze the performance improvement of the surrogate-relevator pairs. Table~\ref{tabPf} shows the $\pi_f$ values achieved by the meta-model with different surrogate-relevator pairs. The meta-model achieves a remarkable performance improvement with an average reduction of up to 77\% of the number of true function evaluations (RF-RF). On individual problems the meta-model achieves up to 94\% performance improvement (SVM-RF).

As was the case for the improvement in speed of convergence, the best performing relevator function is Random Forest closely followed by k-nearest neighbors. It is compelling that the best performing surrogate function on average is a simple regression tree closely followed by SVM. 

Regarding the performance achieved on individual problems, it is worth noting that the SVM relevator is unable to improve the optimization performance on the Repressilator. The performance improvement achieved on the other problems by using the SVM as relevator is on par with others. The Gaussian processes surrogate exhibits the same issue.

\begin{table}[!t]
\caption{Speedup of the optimization obtained with a meta-model instance on the Repressilator/ Metabolic/ S-system problem and the average speedup (bold font).}
\centering
\resizebox{.49\textwidth}{!}{
\begin{tabular}{l|c|c|c|c|c}
\hline
$\frac{\text{\Surr} \rightarrow}{\downarrow \text{\Rel}}$ & TREE & KNN & GP & SVM & RF  \\
\hline
\multirow{2}{*}{KNN} & 0.41/0.83/0.87  & 0.37/0.5/0.78  & \emph{0.00}/0.47/0.81 & 0.81/0.50/0.86  & 0.33/0.24/0.89   \\
& \textbf{0.70} & \textbf{0.56} & \textbf{0.42} & \textbf{0.72} & \textbf{0.49}  \\
\multirow{2}{*}{SVM} & \emph{0.00}/0.72/0.84  & \emph{0.00}/0.59/0.70 & \emph{0.00}/0.43/0.82  & \emph{0.00}/0.56/0.87  & \emph{0.00}/0.62/0.85 \\
& \textbf{0.52} & \textbf{0.43} & \textbf{0.42} &\textbf{0.48} & \textbf{0.49} \\
\multirow{2}{*}{RF}  & 0.73/0.72/0.86  & 0.26/0.65/0.74  & \emph{0.00}/0.61/0.87  & 0.31/0.82/0.94  & 0.36/0.77/0.89  \\
& \textbf{0.77} & \textbf{0.55} & \textbf{0.49} & \textbf{0.69} & \textbf{0.67}
\end{tabular}
}
\label{tabPf}
\end{table}

The best performing surrogate-relevator pairs are heterogeneous. Learning the surrogate function and the relevator function are clearly independent tasks that require different learning methods. Overall, given the results of the empirical evaluation and taking into account the computational time needed by the different learning methods, we recommend choosing a strong and robust learner for the relevator function such as Random Forest. The computational cost can be leveraged by the choice of a simpler learner for the surrogate function without compromising the performance.

\section{Related work}

\begin{figure*}[!ht]
\begin{center}
\includegraphics[width=.9\textwidth]{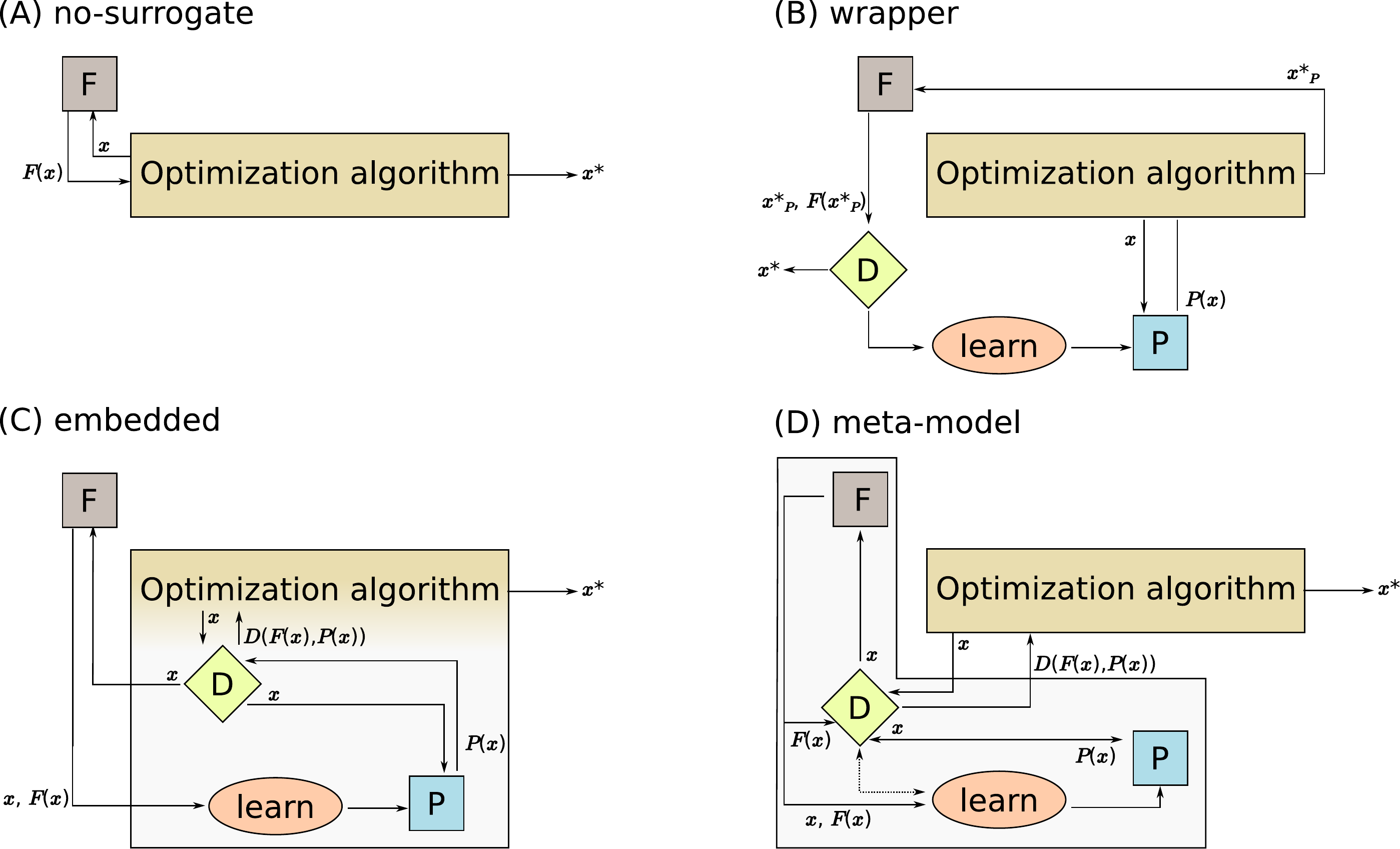}
\end{center}
\caption{Different approaches to surrogate-based optimization. Optimization without surrogates (A), two state-of-the-art classes of surrogate-based optimization methods, wrapper (B) and embedded (C), and the meta-model framework proposed in this paper (D). In the four illustrations, $\Fun$ denotes the objective function, $\Predictor$ the surrogate, and $\Dec$ the decision function corresponding to the substitution strategy. The arrows denote the flow of values between the different components of the optimization method.\label{fig-sa}}
\end{figure*}

In the literature on surrogate-based optimization, the substitution strategy $\Dec$ is referred to as a surrogate management strategy \cite{jin2011}. Figure~\ref{fig-sa} depicts the categorization of the state-of-the-art surrogate-based optimization methods into two classes of wrapper (B) and embedded (C) methods. To better understand the figure, consider first the simple situation of a numerical optimization algorithms that do not use surrogates (A). In such an environment, the optimization method interacts only with the true objective function $\Fun$ by requesting numerous evaluations of different candidate solutions $x \in \fSupp$. At the end, the method reports the optimal solution $x^\ast$ that minimizes (or maximizes) the value of $\Fun$.

Wrapper methods place the surrogate management strategy outside the optimization method. Following this approach, the wrapper first initializes the surrogate $\Predictor$ using a sample of candidate solutions $x \in \fSupp$ and their respective objective evaluations $\Fun(x)$. In consecutive iterations, the wrapper first runs the optimization method using the surrogate model $\Predictor$, obtaining a solution $x_P^\ast$. Next, it evaluates this solution using the true objective function. Finally, the solution $x_P^\ast$ and its evaluation $\Fun(x_P^\ast)$ are added to the training set and the surrogate model is updated (re-learned) before running the next iteration. Note that wrapper methods use a fixed surrogate management strategy that is encoded in the wrapper. Recent developments of the wrapper-approach methods include the methods for constrained numerical optimization COBRA \cite{regis2013} and SOCOBRA \cite{bagherietal2015}.

Embedded methods encode the substitution strategy within the optimization method. Following this approach, the decision on whether to use the surrogate or the true objective function is based on various artifacts of the algorithm \cite{jin2011}. In particular, population-based evolutionary optimization methods use the surrogate model $\Predictor$ to evaluate the offspring candidates for the next generation of individuals. On the other hand, the selection of the top candidates to be actually included in the next generation, is based on the evaluation of the true objective function $\Fun$. A simpler, generation-based management strategy evaluates the surrogate function in some generations, and the true objective function in others. Following the embedded approach, numerous new variants of the classical optimization methods in general \cite{dasetal2016}, and \cite{su2008,rammohan2015} in particular, have been developed.

In sum, the comparison of the wrapper and embedded class of methods, depicted in Figure~\ref{fig-sa}, shows the following. Wrapper approaches are inflexible when it comes to the substitution strategy, since they force the evaluation of the surrogate function within the wrapped optimization method, while the true objective function can only be evaluated from outside the method. On the other hand, embedded approaches are more flexible, but their decision function relies directly on the current state of the core optimization algorithm. Also, they requires re-implementation or modification of an existing implementation of the base optimization method.

The proposed meta-model framework for surrogate-based optimization combines the simplicity of the wrapper approaches with the flexibility of the embedded approaches. On one hand, like in wrapper methods, the meta model can be coupled with any core optimization method since it is used as a black box (see Figure~\ref{fig-sa}(D)). Unlike other wrapper methods, the surrogate model and the substitution strategy are coupled with the true objective function in a manner independent from the optimization algorithm. On the other hand, as in embedded methods, the substitution strategy of the meta model is more flexible. While embedded approaches base the substitution decision on the artifacts of the optimization algorithm, the meta-model substitution strategy dynamically adapts to the solution space of the optimization problem at hand.

\section{Conclusions}

The main contribution of this paper is the novel meta-model framework for surro\-gate-based optimization. In contrast with the prevailing focus of existing surrogate-based optimization methods on learning accurate surrogate models, the proposed framework involves two learning components. One of these learns the surrogate model and the other learns the decision function that takes the decision on when to substitute the true objective function with the surrogate model.

The results of the empirical evaluation of the meta-level framework confirm our initial hypothesis that the selection of appropriate surrogate and decision-function models can have significant influence on the overall performance of surrogate-based optimization. Moreover, the results show that the meta-model performance is more sensitive to the selection of the decision-function model: while almost all learning algorithms (except linear regression) lead to useful surrogate models, only random forests, nearest neighbors and support vector machines are appropriate in the role of decision-function models.

More specific contribution of the paper is a novel surrogate-based approach to estimating the parameters of ordinary differential equations. We consider three parameter estimation tasks with different complexity and showed that, for these tasks, the use of a meta model improves the efficiency of optimization. In particular, the use of the relevator meta model for surrogate-based optimization significantly and efficiently improves the convergence rate and the final result of the optimization, when considering a limited number of evaluations of the true objective function.

The presented meta-model framework significantly contributes to the current machine learning literature by establishing a new paradigm for coupling optimization and machine learning methods. While most of the studies in the machine learning literature are currently based on the sequential model-based optimization (SMBO) paradigm \cite{jones1998}, the proposed framework opens a whole new avenue of research, rich with opportunities for coupling surrogate models with other state-of-the-art optimization methods. The proposed framework is ready to be applied in the context of the currently very active machine learning research on hyper-parameter optimization \cite{wistuba2018}, algorithm configuration \cite{eggensperger2018}, and to other meta-learning tasks \cite{brazdil2018}.

The current conceptualization of the meta-level framework is limited to single-objective, unconstrained optimization. Its generalization towards dealing with multiple objective functions and/or constraints represents two possible directions for further research. Despite the fact that we have applied the framework to numerical optimization only, it is general enough to address also combinatorial or mixed optimization problems. The evaluation of the framework, presented in this paper, is also limited to its coupling with the Differential Evolution method. While this is a typical representative of a more general class of population-based optimization methods, further experimental evaluation is necessary to establish its generality with respect to the selection of the base optimization algorithm. Finally, further evaluation can include comparative analysis of the performance of the meta-model framework relative to the performance of wrapper and embedded surrogate approaches.

\section*{Data availability}
The source code of the implementation of the meta-model framework for surrogate-based parameter estimation, the models and the data used in the experiments are freely available at \url{http://source.ijs.si/zluksic/metamodel}.

\section*{Acknowledgements}
The authors acknowledge the financial support of the Slovenian Research Agency (research core funding No.~P2-0103, No.~P5-0093 and project No.~N2-0128) and the Slovenian Ministry of Education, Science and Sport (agreement No.~C3330-17-529021).

\appendix

\section*{Model of the repressilator}

\begin{align*}
\dot{m_1} &= \alpha_0 + \frac{\alpha}{1 + p_3^n} - dm_1 \\
\dot{m_2} &= \alpha_0 + \frac{\alpha}{1 + p_1^n} - dm_2 \\
\dot{m_3} &= \alpha_0 + \frac{\alpha}{1 + p_2^n} - dm_3 \\
\dot{p_1} &= \beta(m_1 - p_1) \\
\dot{p_2} &= \beta(m_2 - p_2) \\
\dot{p_3} &= \beta(m_3 - p_3)
\end{align*}

\section*{Model of a metabolic NAND gate}

\begin{align*}
\dot{S_3} &= -\frac{S_3Vmax_1}{(S_3 + K_{D1})(1 + \frac{I_1}{K_{I1}})}-\frac{S_3Vmax_2}{(S_3 + K_{D2})(1 + \frac{I_2}{K_{I2}})}+\frac{S_4Vmax_3}{S_4 + K_{D3}}+\frac{S_5Vmax_4}{S_5 + K_{D4}} \\
\dot{S_4} &= \frac{S_3Vmax_1}{(S_3 + K_{D1})(1 + \frac{I_1}{K_{I1}})} - \frac{S_4Vmax_3}{S_4 + K_{D3}} \\
\dot{S_5} &= \frac{S_3Vmax_2}{(S_3 + K_{D2})(1 + \frac{I_2}{K_{I2}})} - \frac{S_5Vmax_4}{S_5 + K_{D4}} \\
\dot{S_6} &= \frac{S_7Vmax_5}{(S_7 + K_{D5})(1 + \frac{S_3}{K_{I3}})} - \frac{S_6Vmax_6}{S_6 + K_{D6}} \\
\dot{S_7} &= \frac{S_6Vmax_6}{S_6 + K_{D6}} - \frac{S_7Vmax_5}{(S_7 + K_{D5})(1 + \frac{S_3}{K_{I3}})}
\end{align*}

\section*{S-system model of a genetic network}

\begin{align*}
\dot{X_1} &= \alpha_1X_3^{g_{13}}X_5^{g_{15}} - \beta_1X_1^{h_{11}} \\
\dot{X_2} &= \alpha_2X_1^{g_{21}} - \beta_2X_2^{h_{22}} \\
\dot{X_3} &= \alpha_3X_2^{g_{32}} - \beta_3X_2^{h_{32}}X_3^{h_{33}} \\
\dot{X_4} &= \alpha_4X_3^{g_{43}}X_5^{g_{45}} - \beta_4X_4^{h_{44}} \\
\dot{X_5} &= \alpha_5X_4^{g_{54}} - \beta_5X_5^{h_{55}} 
\end{align*}

\EOD

\end{document}